\documentclass[twoside]{article}

\usepackage{PRIMEarxiv}

\usepackage[utf8]{inputenc} 
\usepackage[T1]{fontenc}    
\usepackage{hyperref}       
\usepackage{url}            
\usepackage{booktabs}       
\usepackage{amsfonts}       
\usepackage{nicefrac}       
\usepackage{microtype}      
\usepackage{lipsum}
\usepackage{fancyhdr}       
\usepackage{graphicx}       
\graphicspath{{media/}}     
\usepackage{caption}
\usepackage[export]{adjustbox}
\usepackage[table,xcdraw,dvipsnames]{xcolor}
\usepackage{float}
\usepackage{setspace} 
\usepackage{lineno}
\usepackage{rotating}
\usepackage{graphicx}
\usepackage{subfigure}
\usepackage{multirow}
\usepackage{amsmath}
\usepackage{makecell}
\usepackage[flushleft]{threeparttable}

\usepackage{soul}
\usepackage[toc,page]{appendix}

\usepackage{lmodern,textcomp}

\title{Hybrid Machine Learning techniques in the management of harmful algal blooms impact
\thanks{\textit{\underline{Citation}}: 
\textbf{Molares-Ulloa, A., Rivero, D., Ruiz, J. G., Fernandez-Blanco, E., \& de-la-Fuente-Valentín, L. (2023). Hybrid machine learning techniques in the management of harmful algal blooms impact. Computers and Electronics in Agriculture, 211, 107988. DOI:10.1016/j.compag.2023.107988.}} 
}
\author{
  Andres Molares-Ulloa, Daniel Rivero, Enrique Fernandez-Blanco \\
  Universidade da Coruña, Department of Computer Science and Information Technology, \\
  Faculty of Computer Science, 15071, A Coruña, Spain \\
  Centro de investigación CITIC, Department of Computer Science and Information Technology, \\ 
  University of A Coruña, 15071, A Coruña, Spain \\
  \texttt{andres.molares@udc.es} \\
   \And
  Jesús Gil Ruiz, Luis de-la-Fuente-Valentín \\
Universidad Internacional de La Rioja. Avenida de la Paz, 137, 26006 Logroño, La Rioja, Spain, \\
Escuela Superior de Ingeniería y Tecnología \\
}

\begin{document}
\maketitle

\begin{abstract}
Harmful algal blooms (HABs) are episodes of high concentrations of algae that are potentially toxic for human consumption. Mollusc farming can be affected by HABs because, as filter feeders, they can accumulate high concentrations of marine biotoxins in their tissues. To avoid the risk to human consumption, harvesting is prohibited when toxicity is detected. At present, the closure of production areas is based on expert knowledge and the existence of a predictive model would help when conditions are complex and sampling is not possible. Although the concentration of toxin in meat is the method most commonly used by experts in the control of shellfish production areas, it is rarely used as a target by automatic prediction models. This is largely due to the irregularity of the data due to the established sampling programs. As an alternative, the activity status of production areas has been proposed as a target variable based on whether mollusc meat has a toxicity level below or above the legal limit. This new option is the most similar to the actual functioning of the control of shellfish production areas. For this purpose, we have made a comparison between hybrid machine learning models like Neural-Network-Adding Bootstrap (BAGNET) and Discriminative Nearest Neighbor Classification (SVM-KNN) when estimating the state of production areas. The study has been carried out in several estuaries with different levels of complexity in the episodes of algal blooms to demonstrate the generalization capacity of the models in bloom detection. As a result, we could observe that, with an average recall value of 93.41\% and without dropping below 90\% in any of the estuaries, BAGNET outperforms the other models both in terms of results and robustness.
\end{abstract}

\keywords{Machine Learning \and Harmful Algal Blooms \and Biotoxins \and Aquaculture \and  Hybrid Techniques}

\section{Introduction}
\label{S:1}

Bivalve molluscs find their own food source in the microalgae (phytoplankton) present in the aquatic environment. Some of these microalgae belong to species which can highly increase their numbers in a given location. These phenomena are called ``algal blooms''. They are known as ``harmful algal blooms'' (HABs) when caused by microalgal species with harmful effects on human health, the environment, tourism and aquaculture (\cite{BURKHOLDER1998}). These microalgae produce toxins which can be classified into three types depending on the type of poisoning they produce: \textit{Paralytic Shellfish Poisoning} (PSP), \textit{Amestic Shellfish Poisoning} (ASP) or \textit{Diarrhoeic Shellfish Poisoning} (DSP). The latter being the most common in galician coast (northwest Spain)(\cite{Vilas2008ria}). This region is of great importance due to its high production of molluscs (Galicia generates around 40\% of the European mussel production (\cite{apromar})). Although it is possible to detect them practically all year round, their abundance varies seasonally, as well as depending on certain external factors (\cite{Vilas2008ria}). Although these events correspond to natural phenomena known for centuries (\cite{hallegraeff2003harmful}), in the last decades these events seem to have increased in frequency, intensity and geographical distribution (\cite{hallegraeff2003harmful}). There is currently an important scientific interest in understanding the causes and effects of the spatial and temporal distribution of algal species that make up HABs, as their potential effects include ecosystem alterations, public health problems, reduced tourism and social problems, among others, which imply important economic losses (\cite{CORLETT2007245, MARANDA2007632}). For this reason, constant monitoring of these phenomena is necessary to take preventive action when they appear. HABs are a natural phenomenon, and their occurrence cannot be deliberately prevented. Therefore, an active surveillance plan must be maintained to monitor their occurrence and to determine the geographic location of the upwelling.\\
\\
Within the European Union, the management of openings and closures of production areas is carried out by analysing the presence of toxicity in mollusc meat (\cite{ReglCE853/2004}), (\cite{UE786/2013}) and (\cite{UE627/2019}). Where such analyses are not possible, the legislation allows the authorities responsible for such monitoring to make decisions based on endogenous and exogenous factors that favour the proliferation of toxic phytoplankton. At present, the closing of the production areas is based on expert knowledge and lacks predictive models to support it. The occurrence of these events represents serious disruptions and sometimes economic losses for the industry. This drawbacks could be reduced with the existence of predictive models that allow for the creation of contingency plans (\cite{jin2008value}). The use of machine learning techniques is the option that offers the best results (\cite{cruz2021review}). There are two main approaches to the creation of these models. This approach depends on the prediction goal of the ML models. On the one hand, we have models that focus on predicting the concentration of very specific cells such as the \textit{Dinophysis acuminata} (\cite{VeloSurez2007ArtificialNN, aguilar2017prediccion}), or species of the genus \textit{Pseudo-nitzschia} (\cite{vilas2014support}) or \textit{Karlodinium} (\cite{guallar2016artificial}). This approach is more specific but maybe insufficient when predicting HABs composed simultaneously of several species. On the other hand, we have the more generalist approach, which seeks to predict biomarkers that are highly related to HABs, such as chlorophyll-a concentration (\cite{rahman2013algae}) or the toxin concentration itself (\cite{molares2020,MOLARESULLOA2022106956}). Prediction based on the presence of toxin is the least common method, although this information is used by experts in the control of shellfish production areas. This is largely due to the irregularity of the data due to the established sampling programs. As an alternative, the activity status of the production areas has been proposed in this study as an objective variable, depending on whether the meat of cultured molluscs has a toxicity level below or above the legal limit.\\
\\
From a machine learning perspective, there is a trend towards increasing model complexity.  Algorithms such as Support Vector Machine (SVM) have offered good results thanks to their ability to work with a small amount of data and their good generalization capacity (\cite{RIBEIRO200886, 7043865}). However, it has a high computational cost and tends to overfit when applied to high-dimensional multivariate time series. The most recent approaches are based on Random Forest (RF) and Artificial Neural Network (ANN), both for predicting HABs and toxins in molluscs meat. On one hand, RFs provide explicit rules that can be easily interpreted by humans, but present difficulties when dealing with high-dimensional multivariate data (\cite{DEROT2020101906, HARLEY2020101918}). On the other hand, ANNs are better suited to this type of data although they are difficult to interpret (\cite{1553742, GUO2020111731}). In general, other studies work with data sampled at regular time intervals, which allows the creation of complete time series data sets. This is a clear advantage when training ML models that enables the use of methods such as Autoregressive Integrated Moving Average (ARIMA) (\cite{CHEN201558}) or Convolution Neural Network (CNN) (\cite{lai2018modeling}). In cases such as the one we are studying, where the sampling intervals are irregular and the data has large imbalances, the results are worse. The unbalance between positive and negative cases is the reason for this behavior. In addition, HAB episodes correspond to the minority class, making their detection a great challenge. Therefore, the aim of this work is to alleviate these problems and improve the results in the prediction of HABs. According to the current state of the art concerning this type of problem, the best results obtained are based on ensemble techniques such as XGBoost (\cite{rs13193863}). Due to the recent boom of hybrid techniques within the field of artificial intelligence (AI) and its high adaptability capacity (\cite{ruiz1983application}), it has been decided to study the feasibility of applying this type of techniques.\\
\\
In order to support the HABs related opening and closing of production areas, we propose the creation of a predictive model based on hybrid AI techniques. The techniques implemented in this study are: Neural-Network-Adding Bootstrap (BAGNET) (\cite{zhang1999developing}) and Discriminative Nearest Neighbor Classification (SVM-KNN) (\cite{zhang2006svm}). These two methods have not been previously tested in the prediction of HABs. To test the performance of these techniques we have used as a benchmark a set of techniques already applied in other studies related to HAB prediction. In this control group we have used: Random Forest, Artificial Neural Networks (ANN), k-Nearest Neighbour (kNN), Support Vector Machines (SVMs), XGBoost, and Naïve Bayes  (\cite{MOLARESULLOA2022106956}). These models were tested in the literature to support very localized mussel production areas. These regions do not cover all possible situations present in the production areas. This resulted in the creation of local models with low generalization capacity. In contrast, in this work we aim to create models capable of good generalization. For this purpose, the models are trained with data collected from several production areas located in different estuaries. This allows us to test the performance of the models over regions with heterogeneous HABs behavior.\\
\\
The structure of this article is defined as follows: It starts with the definition of the HAB problem and how it affects the seafood industry, as well as the proposed solution. Section \ref{sec:mm} contains a brief explanation of the hybrid techniques used as well as the collection and processing of the dataset is given. The results obtained after the application of these techniques are presented in section \ref{sec:r} and evaluated in section \ref{sec:d} by comparing them with the existing literature. Finally, in section \ref{sec:c}, we present the conclusions drawn from this work and the possible lines of future work that the advances made leave open.

\section{Materials and Methods} 
\label{sec:mm}
\subsection{Dataset and its construction}

Galician coast was chosen for this study because it is one of the main mollusc producing regions in Spain and has a base study to compare the results obtained (\cite{vilas2014support,aguilar2017prediccion,molares2020}). For the creation of the dataset, we used a series of metrics related to HABs and their proliferation collected between the years 2004 and 2019. These data were obtained from oceanographic sampling carried out by the \textit{Instituto Tecnolóxico para o Control do Medio Mariño de Galicia} (INTECMAR) (\cite{intecmar}) in the 42 oceanographic stations distributed along the 5 Galician estuaries (\cite{zonasProduccion}), deciding to eliminate stations recently installed that offer less historical records. Additional data was also collected from \textit{Instituto Español de Oceanografía} (IEO) marnaraia project (\cite{raia}).\\

\begin{figure}[H]
  \centering
  \includegraphics[width=1\textwidth]{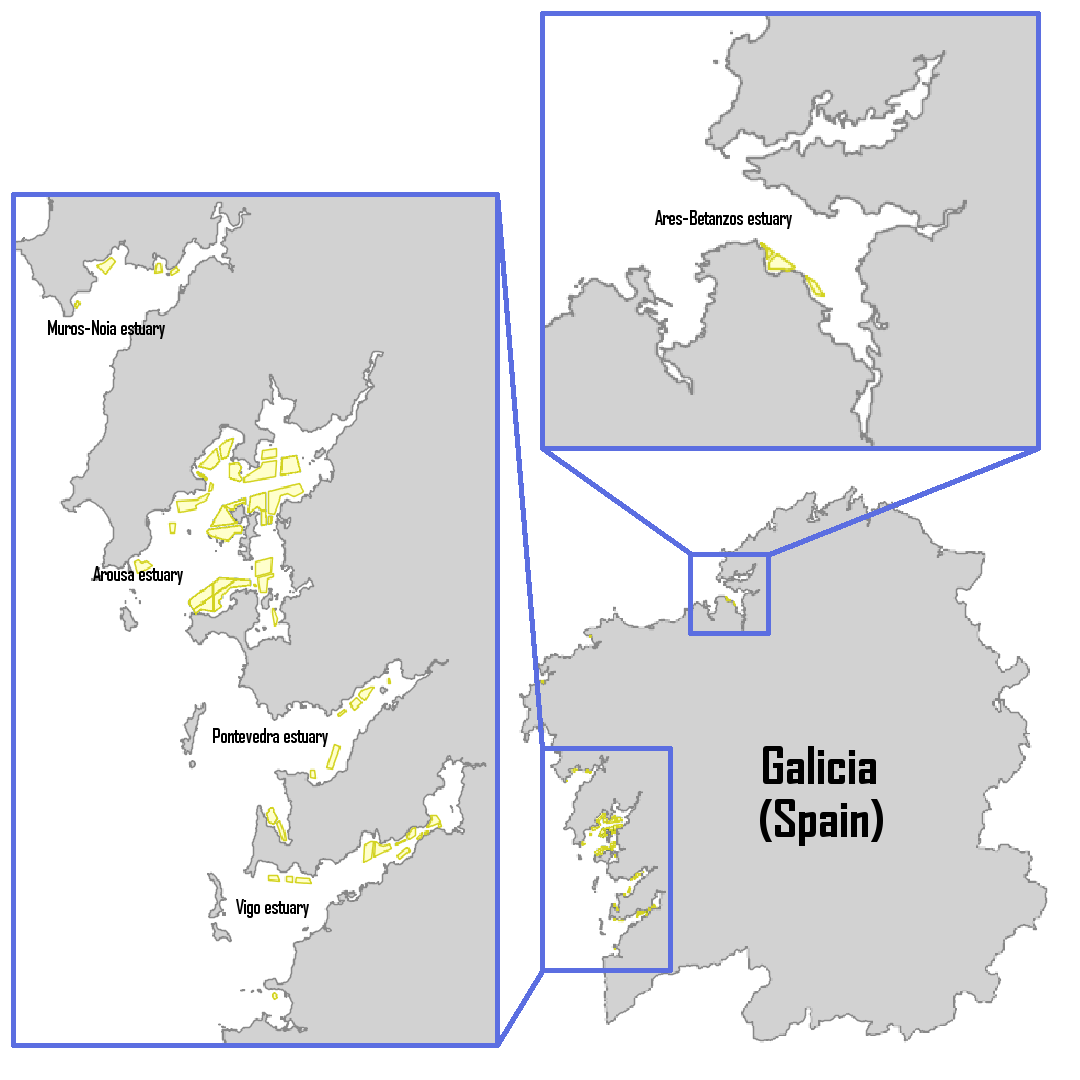}
  \caption{Map of Galicia and the location of the production areas studied marked in yellow.}
  \label{fig:mapa}
\end{figure}

The distribution of these 42 oceanographic stations can be seen in table \ref{tab:distribucion}. In addition, the production areas where the toxicity is analysed are shown for each estuary, being this information the one used to label the samples. The location of these production areas can be seen in Figure \ref{fig:mapa}. Working with several estuaries separately allows us to create study sets with different levels of complexity. The Ares-Betanzos estuary is the simplest with only 4 stations and 2 production zones, while Arousa is the most challenging one with a total of 10 and 22, respectively.

\begin{table}[h]
\centering
\resizebox{1\textwidth}{!}{%
\begin{tabular}{@{}lccccc@{}}
\toprule
\rowcolor[HTML]{ECF4FF} 
\textbf{} &
  \textbf{Ares-Betanzos} &
  \textbf{Muros-Noia} &
  \textbf{Arousa} &
  \textbf{Pontevedra} &
  \textbf{Vigo} \\ \midrule
\textbf{No. oceanographic stations} &
  4 &
  8 &
  10 &
  11 &
  9 \\ 
\textbf{No. production areas} &
  2 &
  4 &
  24 &
  8 &
  12 \\ 
\textbf{No. of samples} & 
    1,564 &
    3,128 & 
    18,768 & 
    6,256 & 
    9,384 \\
\textbf{No. of samples without null values} &
  558 &
  1,440 &
  12,168 &
  3,760 &
  5,112 \\ 
\textbf{Closures without null values} &
  \begin{tabular}[c]{@{}c@{}}193 (35\%)\end{tabular} &
  \begin{tabular}[c]{@{}c@{}}508 (35\%)\end{tabular} &
  \begin{tabular}[c]{@{}c@{}}1903 (16\%)\end{tabular} &
  \begin{tabular}[c]{@{}c@{}}1858 (49\%)\end{tabular} &
  \begin{tabular}[c]{@{}c@{}}1168 (23\%)\end{tabular} \\ 
\textbf{Openings without null values} &
  \begin{tabular}[c]{@{}c@{}}365 (65\%)\end{tabular} &
  \begin{tabular}[c]{@{}c@{}}932 (65\%)\end{tabular} &
  \begin{tabular}[c]{@{}c@{}}10265 (84\%)\end{tabular} &
  \begin{tabular}[c]{@{}c@{}}1902 (51\%)\end{tabular} &
  \begin{tabular}[c]{@{}c@{}}3944 (77\%)\end{tabular} \\ \bottomrule
\end{tabular}%
}
\caption{Distribution of oceanographic stations, production areas and openings and closures in each estuary}
\label{tab:distribucion}
\end{table}

The initial intention was to treat the data as a time series, but after a thorough analysis we found that this was not possible. Most of the data are collected thanks to the INTECMAR monitoring program. In this program, oceanographic stations are sampled on a weekly basis. Therefore, the decision was made to create the dataset samples with the information collected during each week and to label them according to the presence of toxicity on Monday of the following week. This labeling was considered of great interest because the decision to open and close the production areas on Mondays is taken without recent analytical results (the sampling program stops on weekends).\\
\\
It was decided to create independent predictive models for each estuary in order to test the robustness of the models when applied to heterogeneous environments. Taking into account that each Galician estuary has its own oceanographic stations, the number of input features will depend on the estuary where the model will be applied. Each of the 42 oceanographic stations can record the following variables: chlorophyll ``a'', ``b'' and ``c'', \textit{Dinophysis acuminata}, \textit{Dinophysis acuta}, \textit{Dinophysis caudata}, \textit{Dinophysis spp.}, Alexandrium spp., Gymnodinium catenatum, Pseudonitzschia spp., nutrients (phosphate, nitrate, nitrite and ammonium), temperature, salinity and oxygen. Other variables such as the upwelling index and the opening and closing of the production areas were used. These features were processed and adapted to create the datasets in the following way:
    
\begin{itemize}

    \item The maximum values of chlorophyll concentration ``a'', ``b'' and ``c'' have been used.
    
    \item The count of the different phytoplankton cells that produce DSP toxin (\textit{Dinophysis acuminata}, \textit{Dinophysis acuta}, \textit{Dinophysis caudata} and \textit{Dinophysis spp.}) was used.
    
    \item The concentration value of dissolved nutrients (phosphate, nitrate, nitrite and ammonium) was used.
    
    \item The average values of temperature, salinity and oxygen have been used. In addition, with the temperature and salinity values, the absolute difference between the mean of the first 6 meters and the next 6 meters was calculated to detect the presence of thermocline and halocline stratification.
    
    \item We have used the weekly mean value of the upwelling index.
    
    \item Production areas are classified as open or closed depending on whether they have a toxicity level below or above the legal limit, respectively. This classification was used both for the creation of the output parameters and one of the input parameters. The output parameter is composed of the Monday value of the week following the week studied, while the input parameter is composed of the Friday value as this is the day with information sampling closest to the day of the prediction.
    
    \item The production area to which each sample belongs was coded by one hot encoding (\cite{rodriguez2018beyond}).
    
    \item The sampling date was transformed only into the number of the week of the year.
\end{itemize}

After data processing, 5 datasets were obtained (one for each estuary) consisting of the variables reflected in the table \ref{tab:entrada_DSP}. This table is made up of different statistical values like average, maximum or minimum value. In the table we can see how the maximum concentration values of \textit{D. acuminata} vary greatly depending on the estuary. Since this variable is one of the most decisive when estimating closures caused by the presence of DSP toxin, we can foresee that the estuaries with the greatest variability between maximum and minimum values (Arousa and Pontevedra) will be the hardest to make predictions for. These datasets had a large number of inconsistencies in the feature values. This may be due to technical failures, the impossibility to sample or the late creation of certain stations. Since the models only admit that samples with the same dimension, it was necessary to eliminate those that had null values in any of their features. After this filtering, the distribution of openings and closings of the production zones present in each estuary can be seen in the table \ref{tab:distribucion}.

\begin{table}[h]
\centering
\resizebox{1\textwidth}{!}{%
\begin{tabular}{@{}llllllllllll@{}}
\toprule
 & \textbf{Ares-} &  & \textbf{Muros-} &  &  &  & \textbf{Ares-} &  & \textbf{Muros-} &  &  \\
\multirow{-2}{*}{\textbf{Features}} & \textbf{Betanzos} & \multirow{-2}{*}{\textbf{Arousa}} & \textbf{Noia} & \multirow{-2}{*}{\textbf{Pontevedra}} & \multirow{-2}{*}{\textbf{Vigo}} & \multirow{-2}{*}{\textbf{Features}} & \textbf{Betanzos} & \multirow{-2}{*}{\textbf{Arousa}} & \textbf{Noia} & \multirow{-2}{*}{\textbf{Pontevedra}} & \multirow{-2}{*}{\textbf{Vigo}} \\ 
\midrule
\rowcolor[HTML]{D9D9D9} 
 &  &  &  &  &  & Dinophysis acuta &  &  &  &  &  \\
\rowcolor[HTML]{D9D9D9} 
\multirow{-2}{*}{Weekly index}& & & & & & concentration ($Cel*l^{-1}$) & & & & &\\
No. of sampling areas & 1 & 1 & 1 & 1 & 1 & No. of sampling areas & 4 & 10 & 8 & 11 & 9 \\
No. of samples & 1,566 & 18,768 & 3,128 & 6,264 & 9,396 & No. of samples & 1,526 & 17,352 & 2,432 & 5,888 & 8,952 \\
Average value & 26.54 & 26.57 & 26.57 & 26.54 & 26.54 & Average value & 7.71 & 47.36 & 49.61 & 15.49 & 40.32 \\
Maximum value & 53 & 53 & 53 & 53 & 53 & Maximum value & 1,040 & 11,120 & 18,000 & 3,120 & 11,360 \\
Minimum value & 1 & 1 & 1 & 1 & 1 & Minimum value & 0 & 0 & 0 & 0 & 0 \\
\rowcolor[HTML]{D9D9D9} 
Average &  &  &  &  &  & Dinophysis caudata &  &  &  &  &  \\
\rowcolor[HTML]{D9D9D9} 
temperature ($^o$C) &  &  &  &  &  &  concentration ($Cel*l^{-1}$) &  &  &  &  &  \\
No. of sampling areas & 4 & 10 & 8 & 11 & 9 & No. of sampling areas & 4 & 10 & 8 & 11 & 9 \\
No. of samples & 1,308 & 15,408 & 2,096 & 5,192 & 7,896 & No. of samples & 1,526 & 17,352 & 2,432 & 5,888 & 8,952 \\
Average value & 14.66 & 14.16 & 14.59 & 15.45 & 14.73 & Average value & 5.71 & 19.97 & 12.83 & 5.49 & 19.73 \\
Maximum value & 19.34 & 20.10 & 20.00 & 22.68 & 20.18 & Maximum value & 400 & 1,600 & 760 & 360 & 1,440 \\
Minimum value & 10.63 & 11.42 & 11.37 & 10.87 & 11.34 & Minimum value & 0 & 0 & 0 & 0 & 0 \\
\rowcolor[HTML]{D9D9D9} 
 &  &  &  &  &  & Dinophysis spp. &  &  &  &  &  \\
\rowcolor[HTML]{D9D9D9} 
\multirow{-2}{*}{Average salinity} &  &  &  &  &  &  concentration ($Cel*l^{-1}$) &  &  &  &  &  \\
No. of sampling areas & 4 & 10 & 8 & 11 & 9 & No. of sampling areas & 4 & 10 & 8 & 11 & 9 \\
No. of samples & 1,302 & 14,856 & 2,024 & 5,032 & 7,608 & No. of samples & 1,526 & 17,352 & 2,432 & 5,888 & 8,952 \\
Average value & 35.12 & 34.98 & 34.73 & 34.13 & 34.74 & Average value & 36.74 & 5.81 & 4.93 & 4.24 & 9.18 \\
Maximum value & 36.78 & 36.34 & 36.14 & 36.04 & 37.12 & Maximum value & 19,305 & 200 & 280 & 440 & 1,485 \\
Minimum value & 21.12 & 22.46 & 28.08 & 12.51 & 0.56 & Minimum value & 0 & 0 & 0 & 0 & 0 \\
\rowcolor[HTML]{D9D9D9} 
Average &  &  &  &  &  & Ammonium &  &  &  &  &  \\
\rowcolor[HTML]{D9D9D9} 
oxygen ($ml*l^{-1}$) &  &  &  &  &  & dissolved ($\mu mol*l^{-1}$) &  &  &  &  &  \\
No. of sampling areas & 4 & 10 & 8 & 11 & 9 & No. of sampling areas & 4 & 10 & 8 & 11 & 9 \\
No. of samples & 970 & 15,360 & 2,088 & 5,040 & 7,872 & No. of samples & 1,524 & 17,304 & 2,428 & 5,888 & 8,952 \\
Average value & 4.73 & 4.83 & 5.03 & 4.86 & 5.05 & Average value & 1.18 & 0.97 & 1.17 & 2.57 & 1.22 \\
Maximum value & 7.31 & 9.11 & 7.58 & 8.02 & 7.92 & Maximum value & 5.95 & 5.12 & 4.94 & 11.67 & 5.42 \\
Minimum value & 0.16 & 0.07 & 0.07 & 0.08 & 0.08 & Minimum value & 0.14 & 0.05 & 0.20 & 0.03 & 0.05 \\
\rowcolor[HTML]{D9D9D9} 
Thermocline &  &  &  &  &  & Phosphate &  &  &  &  &  \\
\rowcolor[HTML]{D9D9D9} 
 stratification index &  &  &  &  &  & dissolved ($\mu mol*l^{-1}$) &  &  &  &  &  \\
No. of sampling areas & 4 & 10 & 8 & 11 & 9 & No. of sampling areas & 4 & 10 & 8 & 11 & 9 \\
No. of samples & 1,258 & 15,336 & 2,076 & 4,736 & 7,812 & No. of samples & 1,524 & 17,328 & 2,428 & 5,888 & 8,952 \\
Average value & 0.59 & 0.44 & 0.73 & 0.80 & 0.64 & Average value & 0.37 & 0.35 & 0.32 & 0.52 & 0.41 \\
Maximum value & 4.44 & 3.32 & 3.03 & 6.63 & 4.32 & Maximum value & 1.22 & 1.40 & 1.01 & 1.54 & 1.43 \\
Minimum value & 0.00 & 0.00 & 0.01 & 0.00 & 0.00 & Minimum value & 0.06 & 0.02 & 0.04 & 0.04 & 0.03 \\
\rowcolor[HTML]{D9D9D9} 
Halocline &  &  &  &  &  & Nitrate &  &  &  &  &  \\
\rowcolor[HTML]{D9D9D9} 
 stratification index &  &  &  &  &  & dissolved ($\mu mol*l^{-1}$) &  &  &  &  &  \\
No. of sampling areas & 4 & 10 & 8 & 11 & 9 & No. of sampling areas & 4 & 10 & 8 & 11 & 9 \\
No. of samples & 1,222 & 14,832 & 2,012 & 4,576 & 7,560 & No. of samples & 1,524 & 17,328 & 2,428 & 5,888 & 8,952 \\
Average value & 0.71 & 1.26 & 1.12 & 1.21 & 0.49 & Average value & 3.88 & 3.60 & 3.80 & 4.23 & 3.82 \\
Maximum value & 7.53 & 15.01 & 12.69 & 30.67 & 11.26 & Maximum value & 14.85 & 18.36 & 27.51 & 38.12 & 18.39 \\
Minimum value & 0.00 & 0.00 & 0.00 & 0.00 & 0.00 & Minimum value & 0.02 & 0.00 & 0.01 & 0.00 & 0.01 \\
\rowcolor[HTML]{D9D9D9} 
Chlorophyll-a &  &  &  &  &  & Nitrite &  &  &  &  &  \\
\rowcolor[HTML]{D9D9D9} 
 concentration ($mg*l^{-1}$) &  &  &  &  &  & dissolved ($\mu mol*l^{-1}$) &  &  &  &  &  \\
No. of sampling areas & 4 & 10 & 8 & 11 & 9 & No. of sampling areas & 4 & 10 & 8 & 11 & 9 \\
No. of samples & 1,500 & 17,184 & 2,404 & 5,768 & 8,868 & No. of samples & 1,524 & 17,328 & 2,428 & 5,888 & 8,952 \\
Average value & 2.56 & 2.98 & 3.58 & 2.80 & 3.46 & Average value & 0.33 & 0.31 & 0.29 & 0.41 & 0.33 \\
Maximum value & 61.56 & 23.32 & 46.28 & 24.96 & 42.76 & Maximum value & 1.20 & 1.44 & 1.44 & 2.03 & 1.69 \\
Minimum value & 0.04 & 0.04 & 0.06 & 0.02 & 0.07 & Minimum value & 0.03 & 0.01 & 0.02 & 0.01 & 0.01 \\
\rowcolor[HTML]{D9D9D9} 
Chlorophyll-b &  &  &  &  &  & Production area &  &  &  &  &  \\
\rowcolor[HTML]{D9D9D9} 
concentration ($mg*l^{-1}$) &  &  &  &  &  & (One-hot-encoding) &  &  &  &  &  \\
No. of sampling areas & 4 & 10 & 8 & 11 & 9 & No. of sampling areas & 2 & 24 & 4 & 8 & 12 \\
No. of samples & 1,500 & 17,184 & 2,404 & 5,768 & 8,868 & No. of samples & 1,566 & 18,792 & 3,132 & 6,264 & 9,396 \\
Average value & 0.00 & -              0.03 & -              0.03 & -              0.13 & -              0.00 & Average value & 0.50 & 0.04 & 0.25 & 0.13 & 0.08 \\
Maximum value & 3.98 & 0.70 & 2.11 & 0.69 & 5.28 & Maximum value & 1 & 1 & 1 & 1 & 1 \\
Minimum value & -                1.24 & -              1.35 & -              1.73 & -              4.91 & -              1.37 & Minimum value & 0 & 0 & 0 & 0 & 0 \\
\rowcolor[HTML]{D9D9D9} 
Chlorophyll-c &  &  &  &  &  & State of &  &  &  &  &  \\
\rowcolor[HTML]{D9D9D9} 
 concentration ($mg*l^{-1}$) &  &  &  &  &  &  production areas &  &  &  &  &  \\
No. of sampling areas & 4 & 10 & 8 & 11 & 9 & No. of sampling areas & 1 & 1 & 1 & 1 & 1 \\
No. of samples & 1,500 & 17,184 & 2,404 & 5,768 & 8,868 & No. of samples & 1,564 & 18,768 & 3,128 & 6,256 & 9,384 \\
Average value & 0.65 & 0.77 & 0.86 & 0.63 & 0.83 & Average value & 0.30 & 0.17 & 0.35 & 0.50 & 0.27 \\
Maximum value & 23.12 & 7.70 & 9.05 & 11.73 & 9.78 & Maximum value & 1 & 1 & 1 & 1 & 1 \\
Minimum value & -                0.00 & 0.01 & -              0.00 & 0.00 & 0.02 & Minimum value & 0 & 0 & 0 & 0 & 0 \\
\rowcolor[HTML]{D9D9D9} 
Dinophysis acuminata &  &  &  &  &  & Average upwelling &  &  &  &  &  \\
\rowcolor[HTML]{D9D9D9} 
 concentration ($Cel*l^{-1}$) &  &  &  &  &  & index ($m^3*s^{-1}*km^{-1}$) &  &  &  &  &  \\
No. of sampling areas & 4 & 10 & 8 & 11 & 9 & No. of sampling areas & 1 & 1 & 1 & 1 & 1 \\
No. of samples & 1,526 & 17,352 & 2,432 & 5,888 & 8,952 & No. of samples & 1,548 & 18,576 & 3,096 & 6,192 & 9,288 \\
Average value & 217.98 & 260.47 & 236.45 & 312.23 & 283.65 & Average value & 50.59 & 50.59 & 50.59 & 50.59 & 50.59 \\
Maximum value & 8,280 & 23,880 & 8,720 & 43,680 & 12,040 & Maximum value & 2,575.74 & 2,575.74 & 2,575.74 & 2,575.74 & 2,575.74 \\
Minimum value & 0 & 0 & 0 & 0 & 0 & Minimum value & -      4,663.04 & -      4,663.04 & -      4,663.04 & -      4,663.04 & -      4,663.04 \\ \bottomrule
\end{tabular}%
}
\caption{Input values of computational models for DSP toxicity event prediction}
\label{tab:entrada_DSP}
\end{table}

\subsection{Machine Learning Models}

Within the field of machine learning, one of the active areas of study has been the development of hybrid methods that improve classification/prediction performance over approaches based on single learning methods. In general, such methods focus on combining two different machine learning techniques. Based on this idea and previous literature, 2 hybrid machine learning methods such as BAGNET and KNN-SVM have been selected. We have chosen BAGNET inspired by the good performance of bootstrapping based ensemble techniques and the power of Artificial Neural Networks (ANNs). In the case of SVM-KNN, its choice is based on the assumption that the datasets are made up of local data regions with their own distributions. If this were the case, and under the assumption that these local regions were linearly separable, this method could fit the problem very well. These two methods are briefly defined below.

\subsubsection{Neural-Network-Adding Bootstrap}

\begin{figure}[h]
  \centering
  \includegraphics[width=1\textwidth]{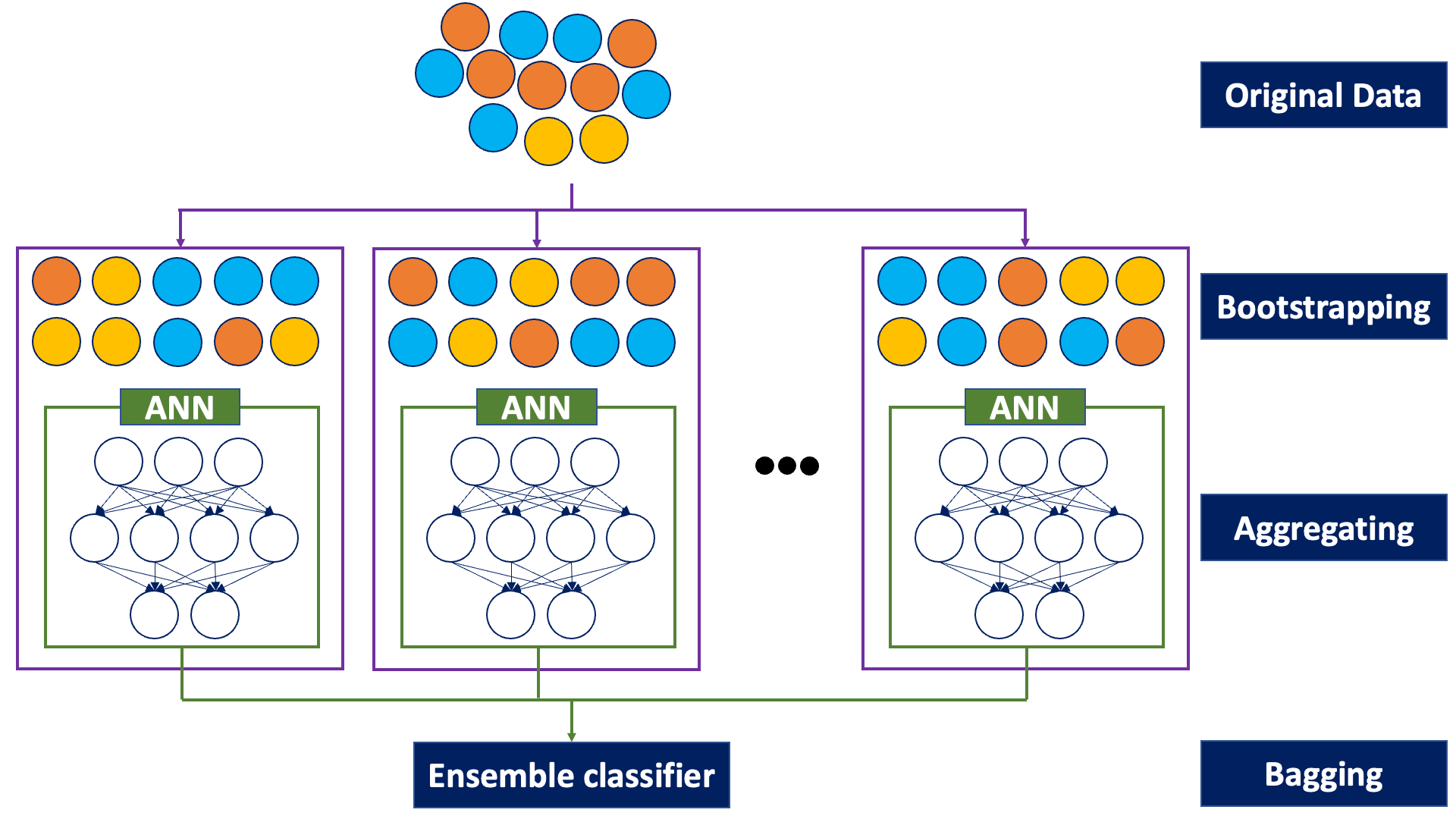}
  \caption{A bootstrap aggregated neural network. This method is based on resampling with replacement of the available dataset and training an individual network on each resampled instance of the original dataset.}
  \label{fig:bagnet}
\end{figure}

The Bagging aggregation method seeks to combine multiple predictors using bootstrap replicates of the training data (\cite{breiman1994bagging}). When this technique is combined with Artificial Neural Networks (\cite{white1992artificial}), the model known as Neural-Network-Adding Bootstrap (BAGNET) emerges. This method is based on resampling with replacement of the available dataset and training an individual network on each resampled instance of the original dataset (\cite{zhang1999developing}). The figure \ref{fig:bagnet} shows a diagram of a BAGNET, in which several neural network models developed to model the same relationship between inputs and outputs are combined together. Instead of selecting a single neural network model, a BAGNET model combines several neural network models to improve the accuracy and robustness of the model. The overall output of a BAGNET is a weighted combination of the individual neural network outputs. Proper determination of the aggregation weights is essential for good modelling performance as is the variant 0.632 bootstrap (\cite{tibshirani1993introduction}). Since the dataset is sampled with replacement, the probability that a given instance is not chosen after $n$ samples is $(1-1/n)^n\approx 0.368$ as $n$ goes to infinity. On the other hand, the probability of being chosen is calculated as $1-(1-1/n)^n\approx e^{-1}\approx 0.632$. This implies that approximately $0.632*n$ unique samples are selected as bootstrap training sets and we would reserve $0.368*n$ out-of-bag samples for testing at each iteration. Therefore, this weighting is calculated by the Eq. \ref{eq:0.632}. Where $ACC_{train}$ is the accuracy computed on the whole training set, and $ACC_{h,i}$ is the accuracy on the out-of-bag sample.

\begin{equation}
   ACC_{boot}=\frac{1}{b}\sum_{i=1}^{b}(0.632\cdot ACC_{h,i}+0.368\cdot ACC_{train})
   \label{eq:0.632}
\end{equation}

\subsubsection{Discriminative Nearest Neighbor Classification}

\begin{figure}[h]
  \centering
  \includegraphics[width=1\textwidth]{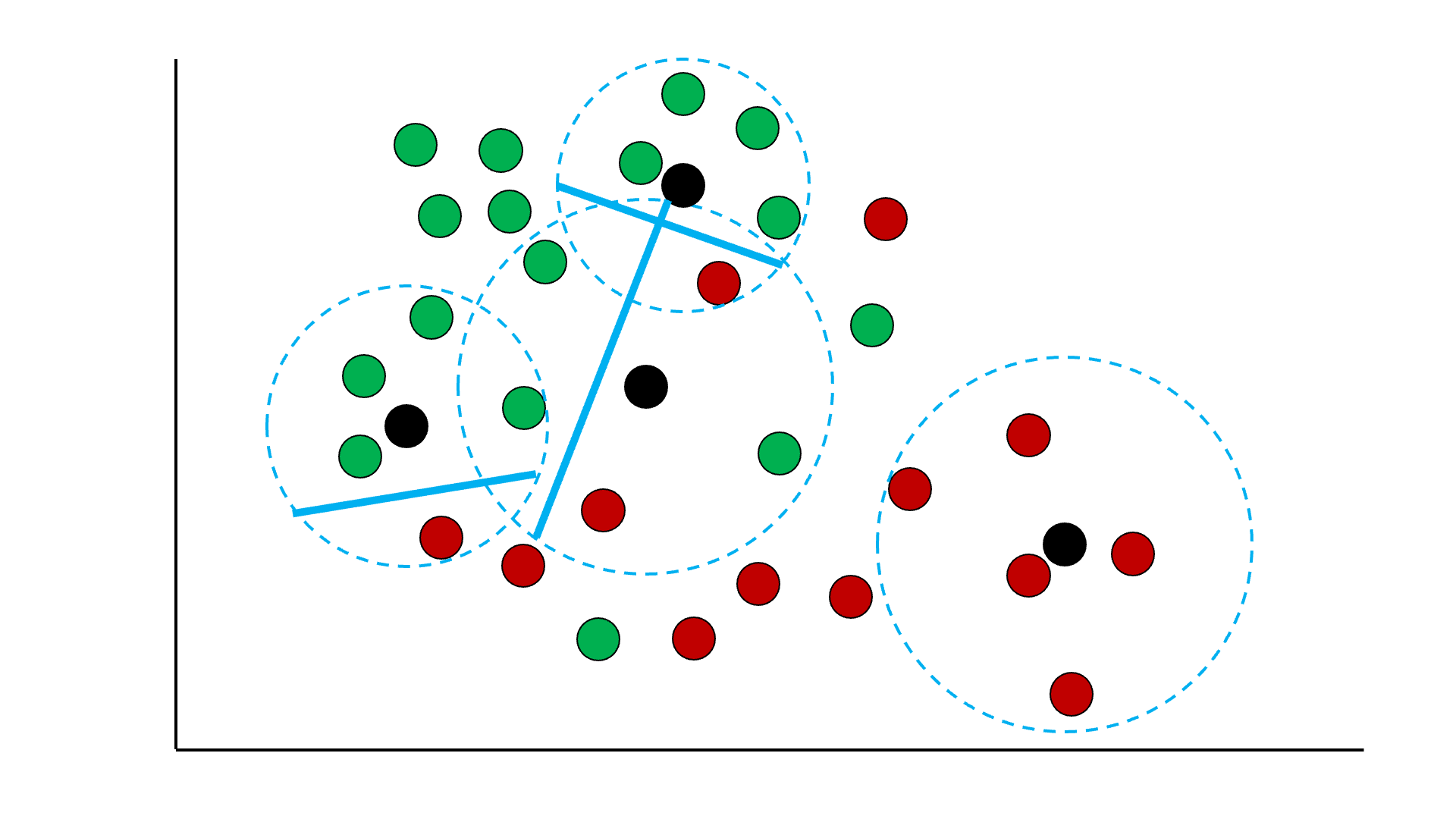}
  \caption{A discriminative nearest nighbor classification. To classify an instance, the kNN algorithm is applied to select the $k$ instances from the training set of closest resemblance. Once these are available, a linear SVM is trained with them, and the instance to be classified is applied to this SVM. Once this is done, the SVM is discarded, since it is of local use and is only valid for that instance.}
  \label{fig:svmknn}
\end{figure}

This system, called SVM-KNN, is based on the hypothesis that a complex separation region can be decomposed into different separated regions that behave linearly locally. To verify or reject this hypothesis, we propose the development of a hybrid model that combines the techniques kNN (\cite{zhang2016introduction}) and SVM (\cite{cortes1995support}). Training an SVM on the entire dataset is slow. However, in the neighborhood of a small number of examples and a small number of classes, SVMs usually perform better than other classification methods (\cite{zhang2006svm}). The philosophy of this method is similar to that of Bottou and Vapnik's ``Local Learning'' (\cite{bottou1992local}), in which they pursued the same general idea using kNN followed by a linear classifier with ridge regularizer. However, by using only an L2 distance, their work was not driven by the constraint of fitting a complex distance function. A diagram of a SVM-KNN is shown in figure \ref{fig:svmknn}, The model would be as follows: to classify an instance, the kNN algorithm is applied to select the $k$ instances from the training set of closest resemblance. Once these are available, a linear SVM is trained with them, and the instance to be classified is applied to this SVM. Once this is done, the SVM is discarded, since it is of local use and is only valid for that instance (\cite{zhang2006svm}).

\subsection{Performance measures}

When analysing the trained models and for subsequent comparison, five statistics were taken into account that were considered relevant for evaluating the results: accuracy, recall, F1-score and kappa coefficient. In the confusion matrix used to calculate the statistics, the closures of the production zones were defined as positive and the openings of the production zones as negative. Thus, True Positives ($TP$) correspond to closures correctly classified as closures, True Negatives ($TN$) identify openings classified as openings, False Positives ($FP$) represent openings misclassified as closures and, finally, False Negatives ($FN$) are closures that have been classified as openings.

Calculated according to Eq. \ref{eq:accuracy} the accuracy estimates the correctness of a binary classification test that identifies or excludes a condition.

\begin{equation}
   Accuracy = \frac{TP+TN}{TP+FP+FN+TN}
   \label{eq:accuracy}
\end{equation}

Due to the risk to the health of the population of introducing molluscs with toxins into the market, we have decided to prioritise a more conservative model in terms of opening production areas. This is reflected in the system by the number of false negatives. A high recall (Eq. \ref{eq:recall}) is related to a lower value of false negatives. For this reason, recall is the reference metric in this study.

\begin{equation}
    Recall = \frac{TP}{TP+FN}
    \label{eq:recall}
\end{equation}

The precision or positive predictive value (Eq. \ref{eq:precision}) denotes the variance of a set of values obtained from repeated measurements of a variable. The smaller the variance, the higher the precision. It is expressed as the ratio of the positive cases well classified by the model to the total number of positive predictions.

\begin{equation}
    Precision = \frac{TP}{TP+FP}
    \label{eq:precision}
\end{equation}

The F1-score (Eq. \ref{eq:f1}) is a widely used metric because it summarises recall and precision in a single metric. Therefore, it is a very useful metric for this study because of the inequality present in the data of closures versus openings.

\begin{equation}
    F1 = \frac{2\cdot (Recall\cdot Precision)}{(Recall + Precision)}
    \label{eq:f1}
\end{equation}

Cohen's kappa coefficient, calculated according to Eq. \ref{eq:kappa}, is a statistical measure that adjusts for the effect of chance on the proportion of observed agreement between two experts. In this equation, $Pr(a)$ represents the observed relative agreement between observers, while $Pr(e)$ is the hypothesised probability of agreement by chance.

\begin{equation}
    K=\frac{Pr(a)-Pr(e)}{1-Pr(e)}
    \label{eq:kappa}
\end{equation}

Once calculated, it is analysed based on the scale of values proposed by Landis and Koch (\cite{landis1977measurement}) (see table \ref{tab:kappa}).

\begin{table}[h]
\centering
\resizebox{0.5\textwidth}{!}{%
\begin{tabular}{@{}cc@{}}
\toprule
\rowcolor[HTML]{ECF4FF} 
\textbf{Kappa statistic} & \textbf{Magnitude of agreement} \\ \midrule
\textless 0.00  & No agreement           \\
0.00-0.20       & Slight                 \\
0.21-0.40       & Fair                   \\
0.41-0.60       & Moderate               \\
0.61-0.80       & Substantial            \\
0.81-1.00       & Almost perfect         \\ \bottomrule
\end{tabular}%
}
\caption{The scale of values proposed by Landis and Koch for interpreting Kappa index values.}
\label{tab:kappa}
\end{table}

Due to the heterogeneous distribution of the data, metrics such as accuracy are not representative of the goodness of fit of the models. Therefore, the combination of the metrics accuracy, F1-score, Kappa index and especially recall will be taken into account due to the importance of avoiding false negatives. This is because classifying a potential closure as an opening of the production area could pose a risk to public health.

\subsection{Experimentation setup}

To ensure the robustness of the models, we have applied the K-fold cross-validation strategy (\cite{wong2015performance}). In this particular study, the K-fold cross-validation strategy selected will be the 10-fold strategy, where $k$ takes value 10. Taking into account the imbalance present in the data, we have used a stratified k-fold. This method ensures that each of the folds will have a balanced amount of data.\\
\\
The following models have been used as benchmarks for comparison: Random Forest, Artificial Neural Networks (ANNs), k-Nearest Neighbour (kNN), Support Vector Machines (SVMs), XGBoost, and Naïve Bayes. Their hyperparameters have been obtained by replicating the experimentation process established in previous studies (\cite{MOLARESULLOA2022106956}) to the new datasets.\\
\\
In the case of the models proposed in this paper, the attributes of the models used were adjusted by a grid search. Grid search is a tuning technique that attempts to compute the optimum values of hyperparameters. It is an exhaustive search that tests all combinations within the values given to the hyperparameters of a model. As described above, BAGNET is an ensemble model composed of a set of networks of neurons. In our particular case, we have worked with an ensemble composed of 50 networks. When testing the operation of BAGNET we have tried different configurations of the networks that compose it, both in the number of layers and in the number of neurons that compose them. The number of neurons used is based on the use of multiples of 2 and combinations of them. The configuration of the individual networks that compose the ensemble is based on the ANNs used in previous studies (\cite{MOLARESULLOA2022106956}). On these values we have followed an empirical experimentation in order to determine the values used in the grid search. The specific configuration of this model is shown in table \ref{tab:parametros_bagnet}. 

\begin{table}[h]
\centering
\resizebox{1\textwidth}{!}{%
\begin{tabular}{@{}lc@{}}
\toprule
\rowcolor[HTML]{ECF4FF} 
\textbf{Neural-Network-Adding Bootstrap} &                                  \\ \midrule
Number of networks                            & 50                               \\
Number of neurons in a one hidden layer network                    & 2, 4, 8, 10, 16, 24 and 32                                \\
\multirow{2}{*}{Number of neurons in a two hidden layers network}                    & {(}4,2{)}, {(}8,4{)}, {(}16,8{)}, {(}24,16{)}, {(}32,16{)}, {(}32, 24{)}, {(}64, 32{)},\\
                    &  {(}128, 32{)}, {(}128,64{)}, {(}192, 128{)} and {(}256, 192{)}       \\
Activation function                      & Sigmoid                         \\
Optimizer                                & Adagrad                          \\
Learning rate                              & 0,05                             \\
Class weight                 & Balanced                               \\ \bottomrule
\end{tabular}%
}
\caption{BAGNET model configuration parameters}
\label{tab:parametros_bagnet}
\end{table}

In the case of the SVM-KNN model, we have tested various values of the parameter $k$, in order to check whether the problem can be effectively decomposed into linearly separable regions. For the choice of the parameter $k$ we have applied the greed search method on the values proposed in the work of Hao Zhang et al. (\cite{zhang2006svm}). We have limited the maximum value of $k$ to 45 as we did not obtain any improvement by increasing the complexity of the subspaces. In addition to what was proposed in said work, we have decided to test the behaviour of the model by varying the values of $C$. The parameters used in the SVM-KNN model can be found in table \ref{tab:parametros_svmknn}.

\begin{table}[h]
\centering
\resizebox{1\textwidth}{!}{%
\begin{tabular}{@{}lc@{}}
\toprule
\rowcolor[HTML]{ECF4FF} 
\textbf{Discriminative Nearest Neighbor Classification} &                                   \\ \midrule
k value                                                & 3, 5, 7, 9, 10, 15, 20, 25, 30, 35, 40, 45, and 50 \\
Minkowski metric                                        & Manhattan distance            \\
Kernel type                                            & Linear                            \\
C value                                                & 0.1, 0.5, 1 and 10                                 \\ \bottomrule
\end{tabular}%
}
\caption{SVM-KNN model configuration parameters}
\label{tab:parametros_svmknn}
\end{table}

\section{Results}
\label{sec:r}

    Using the k-fold cross-validation strategy with $k=10$ yields 10 values for each statistic. To avoid choosing models with good means, but with high variability and low robustness, it was decided to add the standard deviation of each statistic studied as a parameter to be taken into account. The implemented models obtained results that will be shown in the table \ref{tab:results_x_estuary}, consisting of: the estuary where the models have been applied, the trained algorithm, the value of the adjustment parameters (values of $k$ and $C$ for SVM-KNN and the number of neurons per layer for BAGNET), the mean and standard deviation of the metrics studied. As one of the main objectives of this study is to study the generalization capacity of the implemented algorithms, table \ref{tab:resultados} shows a summary of the average behavior of the models in the 5 different estuaries. In turn, the metrics associated with each estuary are determined by the averaged value resulting from applying 10-fold cross-validation. The metrics shown for the SVM-KNN and BAGNET models are those obtained with the hyperparameter configuration that provided the best F1-score values. The best configuration for SVM-KNN was $k=5$ and $C=1$ and for BAGNET was two layers with 192 and 128 neurons respectively (these specific configurations will be noted as SVM-KNN* and BAGNET* respectively). In the table \ref{tab:resultados} we can see how the accuracy values hardly vary depending on the model. If we look at the recall values we can see that BAGNET offers the best results, followed by the models based on ANN, Random Forest and SVM-KNN. If we consider the F1-score values, the best models are those based on Random forest, SVM-KNN and BAGNET, the first one standing out for its low standard deviation. With the exception of the Naïve Bayes-based model, the rest of the models have obtained Kappa values above 0.80. The study of the models applied to each estuary independently allows us to understand how the models behave according to its complexity. These performance allows us to understand why the models have such a high standard deviation, reaching values of more than $\pm$15\% in the recall of the SVM, XGBoost and Naïve Bayes models. This is because the behaviour of the models varies a lot between folds depending on the estuary where they are applied, but vary little between folds in the same estuary. So it can be seen that the models give robust results.

\begin{table}[h]
\centering
\resizebox{1\textwidth}{!}{%
\begin{tabular}{@{}clllllllll@{}}
\toprule

\rowcolor[HTML]{ECF4FF} 
{\color[HTML]{000000} \textbf{ESTUARY}} &
{\color[HTML]{000000} \textbf{MODEL}} &
  \multicolumn{2}{c}{\cellcolor[HTML]{ECF4FF}{\color[HTML]{000000} \textbf{ACCURACY}}} &
  \multicolumn{2}{c}{\cellcolor[HTML]{ECF4FF}{\color[HTML]{000000} \textbf{RECALL}}} &
  \multicolumn{2}{c}{\cellcolor[HTML]{ECF4FF}{\color[HTML]{000000} \textbf{F1-SCORE}}} &
  \multicolumn{2}{c}{\cellcolor[HTML]{ECF4FF}{\color[HTML]{000000} \textbf{KAPPA}}} \\ \midrule
 \parbox[t]{2mm}{\multirow{8}{*}{\rotatebox[origin=c]{90}{ARES-BETANZOS}}} & RF & 92.67 & $\pm$ 1.36 & 86.43 & $\pm$ 4.31 & 87.96 & $\pm$ 2.29 & 82.70 & $\pm$ 3.23 \\
 & ANN           & 94.56 & $\pm$ 1.54 & 93.32 & $\pm$ 4.23 & 91.40 & $\pm$ 2.47 & 85.77 & $\pm$ 3.69 \\
 & kNN           & 93.27 & $\pm$ 1.95 & 87.98 & $\pm$ 6.97 & 88.90 & $\pm$ 3.59 & 84.08 & $\pm$ 4.89 \\
 & \textbf{SVM}           & \textbf{94.83} & \textbf{$\pm$ 1.21} & \textbf{93.82} & \textbf{$\pm$ 3.51} & \textbf{91.84} & \textbf{$\pm$ 1.86} & \textbf{88.07} & \textbf{$\pm$ 2.75} \\
 & XGBoost       & 93.87 & $\pm$ 2.44 & 91.09 & $\pm$ 5.71 & 90.18 & $\pm$ 4.02 & 85.73 & $\pm$ 5.76 \\
 & Naïve Bayes   & 86.41 & $\pm$ 4.36 & 86.77 & $\pm$ 8.93 & 79.80 & $\pm$ 6.41 & 69.67 & $\pm$ 9.63 \\
 & SVM-KNN*      & 93.75 & $\pm$ 1.69 & 89.94 & $\pm$ 6.46 & 89.83 & $\pm$ 3.03 & 85.33 & $\pm$ 4.17 \\
 & BAGNET*       & 94.35 & $\pm$ 1.86 & 93.82 & $\pm$ 3.90 & 91.19 & $\pm$ 2.85 & 87.04 & $\pm$ 4.22 \\ \midrule
 \parbox[t]{2mm}{\multirow{8}{*}{\rotatebox[origin=c]{90}{MUROS-NOIA}}} & RF & 94.80 & $\pm$ 1.43 & 93.09 & $\pm$ 3.32 & 92.46 & $\pm$ 2.05 & 88.49 & $\pm$ 3.14 \\
 & ANN           & 96.28 & $\pm$ 0.97 & 96.53 & $\pm$ 1.40 & 94.70 & $\pm$ 1.30 & 91.45 & $\pm$ 2.31 \\
 & kNN           & 90.78 & $\pm$ 0.91 & 87.65 & $\pm$ 4.55 & 86.67 & $\pm$ 1.46 & 79.63 & $\pm$ 2.08 \\
 & \textbf{SVM}           & \textbf{96.45} & \textbf{$\pm$ 1.03} & \textbf{96.67} & \textbf{$\pm$ 1.24} & \textbf{94.93} & \textbf{$\pm$ 1.38} & \textbf{92.20} & \textbf{$\pm$ 2.19} \\
 & XGBoost       & 96.23 & $\pm$ 0.98 & 96.42 & $\pm$ 1.60 & 94.63 & $\pm$ 1.32 & 91.73 & $\pm$ 2.09 \\
 & Naïve Bayes   & 80.80 & $\pm$ 1.32 & 56.67 & $\pm$ 4.40 & 66.84 & $\pm$ 2.79 & 53.97 & $\pm$ 3.34 \\
 & SVM-KNN*      & 94.75 & $\pm$ 1.19 & 91.36 & $\pm$ 1.91 & 92.29 & $\pm$ 1.64 & 88.32 & $\pm$ 2.57 \\
 & BAGNET*       & 95.98 & $\pm$ 0.95 & 96.54 & $\pm$ 1.64 & 94.30 & $\pm$ 1.26 & 91.20 & $\pm$ 2.01 \\ \midrule
 \parbox[t]{2mm}{\multirow{8}{*}{\rotatebox[origin=c]{90}{AROUSA}}} & RF & 93.62 & $\pm$ 0.51 & 76.78 & $\pm$ 2.43 & 79.28 & $\pm$ 1.78 & 75.52 & $\pm$ 2.07 \\
 & ANN           & 85.09 & $\pm$ 0.88 & 84.31 & $\pm$ 2.64 & 64.29 & $\pm$ 1.81 & 55.31 & $\pm$ 2.27 \\
 & kNN           & 93.11 & $\pm$ 0.46 & 79.58 & $\pm$ 1.89 & 78.61 & $\pm$ 1.48 & 74.51 & $\pm$ 1.74 \\
 & SVM           & 91.09 & $\pm$ 0.56 & 49.71 & $\pm$ 2.94 & 63.94 & $\pm$ 2.67 & 59.34 & $\pm$ 2.89 \\
 & XGBoost       & 90.71 & $\pm$ 0.65 & 54.52 & $\pm$ 2.75 & 65.12 & $\pm$ 2.46 & 60.00 & $\pm$ 2.78 \\
 & Naïve Bayes   & 85.62 & $\pm$ 0.66 & 47.40 & $\pm$ 2.35 & 51.20 & $\pm$ 2.11 & 42.83 & $\pm$ 2.46 \\
 & SVM-KNN*      & 92.23 & $\pm$ 0.47 & 79.62 & $\pm$ 2.31 & 76.54 & $\pm$ 1.57 & 71.89 & $\pm$ 1.84 \\
 & \textbf{BAGNET*}       & \textbf{91.12} & \textbf{$\pm$ 0.76} & \textbf{90.46} & \textbf{$\pm$ 2.61} & \textbf{76.45} & \textbf{$\pm$ 1.74} & \textbf{71.14} & \textbf{$\pm$ 2.18} \\ \midrule
 \parbox[t]{2mm}{\multirow{8}{*}{\rotatebox[origin=c]{90}{PONTEVEDRA}}} & \textbf{RF} & \textbf{92.35} & \textbf{$\pm$ 0.98} & \textbf{93.56} & \textbf{$\pm$ 0.99} & \textbf{92.35} & \textbf{$\pm$ 0.96} & \textbf{84.69} & \textbf{$\pm$ 1.96} \\
 & ANN           & 83.43 & $\pm$ 1.58 & 80.51 & $\pm$ 2.20 & 82.74 & $\pm$ 1.68 & 66.50 & $\pm$ 3.07 \\
 & kNN           & 91.36 & $\pm$ 1.02 & 91.38 & $\pm$ 1.51 & 91.26 & $\pm$ 1.04 & 82.72 & $\pm$ 2.04 \\
 & SVM           & 85.12 & $\pm$ 1.49 & 85.25 & $\pm$ 2.50 & 84.97 & $\pm$ 1.59 & 70.24 & $\pm$ 2.99 \\
 & XGBoost       & 83.56 & $\pm$ 1.33 & 78.06 & $\pm$ 1.67 & 82.42 & $\pm$ 1.38 & 67.07 & $\pm$ 2.67 \\
 & Naïve Bayes   & 72.36 & $\pm$ 0.99 & 52.05 & $\pm$ 1.50 & 65.02 & $\pm$ 1.13 & 44.43 & $\pm$ 1.97 \\
 & SVM-KNN*      & 90.30 & $\pm$ 1.21 & 91.68 & $\pm$ 1.34 & 90.33 & $\pm$ 1.16 & 80.61 & $\pm$ 2.42 \\
 & BAGNET*       & 89.67 & $\pm$ 1.39 & 90.33 & $\pm$ 1.89 & 89.62 & $\pm$ 1.36 & 79.34 & $\pm$ 2.78 \\ \midrule
 \parbox[t]{2mm}{\multirow{8}{*}{\rotatebox[origin=c]{90}{VIGO}}} & RF & 97.31 & $\pm$ 0.66 & 93.87 & $\pm$ 2.23 & 94.10 & $\pm$ 1.50 & 92.36 & $\pm$ 1.93 \\
 & ANN           & 97.42 & $\pm$ 0.63 & 94.97 & $\pm$ 1.69 & 94.41 & $\pm$ 1.37 & 92.24 & $\pm$ 1.83 \\
 & kNN           & 95.37 & $\pm$ 1.10 & 87.41 & $\pm$ 3.90 & 89.60 & $\pm$ 2.62 & 86.62 & $\pm$ 3.31 \\
 & SVM           & 97.35 & $\pm$ 0.66 & 94.55 & $\pm$ 1.65 & 94.23 & $\pm$ 1.44 & 92.51 & $\pm$ 1.87 \\
 & XGBoost       & 97.33 & $\pm$ 0.59 & 93.96 & $\pm$ 1.71 & 94.16 & $\pm$ 1.30 & 92.43 & $\pm$ 1.68 \\
 & Naïve Bayes   & 89.71 & $\pm$ 0.98 & 92.54 & $\pm$ 2.93 & 80.46 & $\pm$ 1.72 & 73.64 & $\pm$ 2.35 \\
 & SVM-KNN*      & 96.72 & $\pm$ 0.68 & 91.36 & $\pm$ 2.64 & 92.70 & $\pm$ 1.58 & 90.59 & $\pm$ 2.01 \\
 & \textbf{BAGNET*}       & \textbf{97.60} & \textbf{$\pm$ 0.67} & \textbf{95.89} & \textbf{$\pm$ 1.62} & \textbf{94.82} & \textbf{$\pm$ 1.44} & \textbf{93.25} & \textbf{$\pm$ 1.87} \\ \bottomrule
\end{tabular}%
}
\caption{Table with metrics for each model itemized by estuary}
\label{tab:results_x_estuary}
\end{table}

\begin{table}[h]
\resizebox{1\textwidth}{!}{%
\begin{tabular}{@{}lllllllllc@{}}
\toprule
\rowcolor[HTML]{ECF4FF} \textbf{MODEL} &
\multicolumn{2}{c}{{\cellcolor[HTML]{ECF4FF} \textbf{ACCURACY}}} &
\multicolumn{2}{c}{{\cellcolor[HTML]{ECF4FF} \textbf{RECALL}}} &
\multicolumn{2}{c}{{\cellcolor[HTML]{ECF4FF} \textbf{F1-SCORE}}} &
\multicolumn{2}{c}{{\cellcolor[HTML]{ECF4FF} \textbf{KAPPA}}} &
\textbf{SOURCE} \\ \midrule
RF & 94.15 & $\pm$2.08 & 88.75 & $\pm$7.19 & 89.23 & $\pm$5.66 & 84.75 & $\pm$6.22 & (\cite{HARLEY2020101918})\\
ANN           & 91.36 & $\pm$6.01 & 89.93 & $\pm$6.86 & 85.51 & $\pm$11.59 & 78.26 & $\pm$15.02 & (\cite{GUO2020111731})\\
kNN           & 92.78 & $\pm$2.01 & 86.80 & $\pm$5.76 & 87.01 & $\pm$4.98 & 81.51 & $\pm$5.16 & (\cite{MOLARESULLOA2022106956})\\
SVM           & 92.97 & $\pm$4.59 & 84.00 & $\pm$17.76 & 85.98 & $\pm$11.72 & 80.47 & $\pm$13.60 & (\cite{7043865})\\
XGBoost       & 92.34 & $\pm$5.13 & 82.81 & $\pm$15.81 & 85.30 & $\pm$11.25 & 79.39 & $\pm$13.75 & (\cite{rs13193863})\\
Naïve Bayes   & 82.98 & $\pm$6.40 & 67.09 & $\pm$19.35 & 68.66 & $\pm$11.33 & 56.91 & $\pm$13.60 & (\cite{MOLARESULLOA2022106956})\\ \midrule
SVM-KNN*      & 93.55 & $\pm$2.46 & 88.79 & $\pm$5.77 & 88.34 & $\pm$6.30 & 83.35 & $\pm$7.17 & \multicolumn{1}{c}{}\\
\textbf{BAGNET*}       & \textbf{93.75} & \textbf{$\pm$ 3.20} & \textbf{93.41} & \textbf{$\pm$ 3.61} & \textbf{89.27} & \textbf{$\pm$ 6.94} & \textbf{84.40} & \textbf{$\pm$ 8.61} & \multicolumn{1}{c}{\multirow{-2}{*}{Proposed here}}\\ \bottomrule
\end{tabular}%
}
\caption{Summary table with the metrics of each model averaged in the 5 estuaries.}
\label{tab:resultados}
\end{table}

\begin{figure}[h]
  \centering
  \includegraphics[width=1\textwidth]{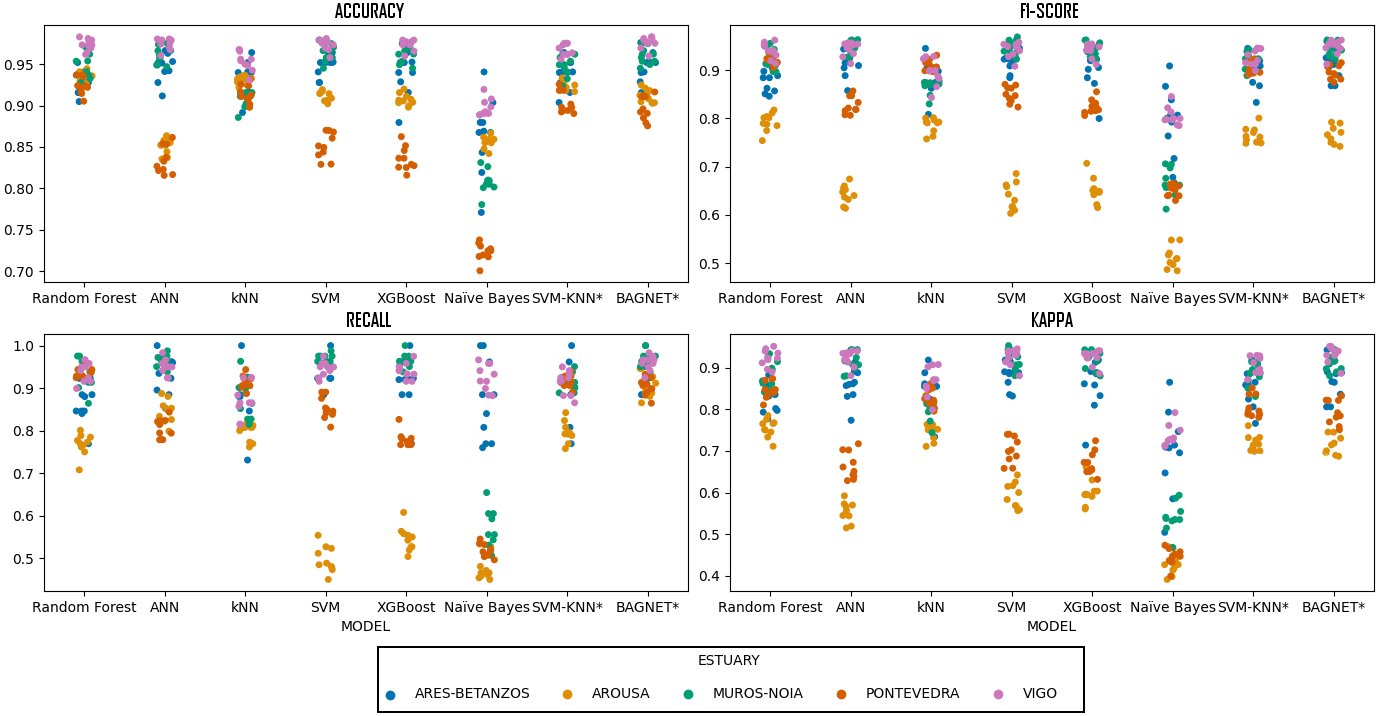}
  \caption{Comparison of accuracy, recall, f1-score and kappa between benchmark models and the best results of SVM-KNN and BAGNET for the prediction of HAB episodes by DSP.}
  \label{fig:metrics}
\end{figure}

In figure \ref{fig:metrics} we can see how the models perform depending on the estuary where they are applied. When separating the metrics obtained according to the production area where they are applied, we can see how the models have less difficulty in making predictions in estuaries such as Ares-Betanzos, Muros-Noia and Vigo. The worst results are obtained in the Pontevedra estuary and the Arousa estuary.

\section{Discussion} 
\label{sec:d}

As we can see in the table \ref{tab:resultados}, very good results of around 92\% of accuracy are obtained. These values are very similar regardless of the model applied. This is due to the fact that the datasets present a large imbalance between positive and negative cases (table \ref{tab:distribucion}). Because of this, our comparison focuses on the recall value obtained. Taking this metric into account, BAGNET is positioned in first place, reaching recall values of 93.41\%, followed by ANN, SVM-KNN and Random Forest, all with a recall rate of around 89\% and a higher standard deviation in their results. These models, with the exception of ANN, reached a Kappa index value above 0.8 when compared to the real values. This is an ``almost perfect'' agreement according to the scale of values proposed by Landis and Koch (\cite{landis1977measurement}). It should be noted that although BAGNET is the model that offers the best recall values, it is a model that offers somewhat lower F1-score values than its alternatives. This is because, like the ANN model, they are configured to weight the loss function (during training only). This can be useful to tell the model to ``pay more attention'' to samples from an under-represented class. This results in an increase in false positives in exchange for being the models with the lowest number of false negatives, the latter being of great importance as explained above.\\
\\
It should be noted that there is no reference dataset on which to test the effectiveness of the models. This is due to the fact that the data used for their construction is provided ad-hoc for the respective studies. Furthermore, it is noteworthy that there are very few studies that aim to predict the presence of toxin in mussel meat (biomarker used by the EU for the management of shellfish production areas). As mentioned earlier in the introduction to this study, most studies determine bloom/no boom based on the concentration of certain phytoplankton species present in the environment. This situation increases the complexity when making comparisons within this field of study. In order to make as fair a comparison as possible between our models and the most common state-of-the-art models, they have all been applied to the same dataset defined in this study.(table \ref{tab:resultados}).\\
\\
Among the studies that do work with the presence of toxin in mussels, those applied to very localized areas stand out (\cite{MOLARESULLOA2022106956}). In this situation, the kNN technique was the most effective, with recall values of 97\%. But when we applied this technique to our data, which were collected over larger regions and contained more variability, we found that the effectiveness of this model decreased in favor of ensemble-based alternatives. kNN offers recall values of 86\% compared to 93\% for the BAGNET model. The proposed BAGNET and SVM-KNN models are a better options for larger environments. Other works, such as Harley et al. (2020) (\cite{HARLEY2020101918}), also work on the prediction of toxins present in mussels. Although in their case the type of toxin studied is PSP, it has been considered relevant to highlight this due to the small number of papers presenting this approach. They have achieved recall values of 81\% using Random Forest for PSP prediction. As can be seen in the table \ref{tab:resultados}, applying Random Forest in the prediction of DSP blooms improves the results compared to the other type of toxin, with a recall of 88\%. Although these results are still lower than those obtained with BAGNET and SVM-KNN.\\
\\
One of our goals is to obtain a model that is extrapolable to other regions with the same problems and that can also be adjusted and provide good results. For this reason, we have applied the predictive models to several estuaries composed of multiple production areas. In doing so, we have found that the models do not behave the same regardless of the region in which they are applied (figure \ref{fig:metrics}). If we analyse the reason why the models offer worse results in estuaries such as Arousa, this is mainly due to two factors. We have to take into account that this is the estuary studied with the greatest imbalance in the labels of its dataset (16\% positives - 84\% negatives), which makes it a challenge for the models to be able to generalise with a smaller number of cases. Moreover, this estuary is the largest we have studied. Composed of a total of 24 mussel production areas in floating hatcheries located along the entire estuary, it represents a greater variability of events that the models have to deal with. Despite these setbacks, we can see that the BAGNET model still offers recall values of around 90.46\% in the Ría de Arousa. The other model that obtains recall values higher than 80\% in this estuary is ANN with 84.31\%, but this model achieves 10\% less F1-score and 15\% less Kappa index than BAGNET.

\section{Conclusions} 
\label{sec:c}
Hybrid techniques of machine learning algorithms obtain better results than the simple techniques studied in the literature. We also verified how the BAGNET method, with an average recall value of 93.41\% and without falling below 90\% in any of the estuaries, exceeds the results obtained by other hybrid methods proposed in the literature, such as XGBoost or Random Forest. This positions BAGNET as the best algorithm for DSP control of HABs when sampling conditions are unfavorable.\\
\\
To make this comparison, we have had to implement the models proposed in other studies. This is due to the fact that, to date, there is no reference data set in the prediction of HAB that allows an objective comparison of the models studied in other works. This is because the data sets used are provided ad-hoc for each study. For this reason, the creation of a complete and public data set that can be used for future comparisons has been considered of great importance.\\
\\
In order to create more complex test environments and increase the generalization capacity of the system, we have created data sets grouping all the production areas of each estuary. Although this process carries the risk of increasing the complexity of the system, the use of ensemble-based techniques and bootstrapping, such as BAGNET, has allowed us to address this problem with very good results.\\
\\
In the course of this study, one of the main problems we faced was the large number of gaps in the time series data. This is an important limitation when working with the data and interpreting it. Being able to create a data set consisting of a complete time series would allow new approaches to HAB prediction.\\
\\
The importance of having a system capable of monitoring and predicting the appearance of HAB phenomena is well known. Poor planning before the formation of these natural phenomena can cause significant economic losses in the fishing and shellfish businesses. In addition, and more importantly, there is the risk of introducing toxin-containing shellfish onto the market due to failure to close the production areas on time due to poor planning. This risk poses a significant danger to the health of the population. For all these reasons, we believe that models such as the one proposed can be of great help in addressing this problem. In particular, it is worth highlighting factors such as the creation of models focused on the control of shellfish production areas and the creation of models with greater generalization capacity, capable of being applied in multiple regions. These last factors are where our work stands out, improving the results obtained in previous studies.


\section*{Acknowledgments}
The authors want to acknowledge the support from INTECMAR, who has provided most of the data for this work; and CESGA, who allowed the conduction of the tests on their installations. Funding for open access charge: Universidade da Coruña/CISUG.

\section*{Data availability and reproducibility}
The source code of the analysis is available on GitHub: \url{https://github.com/AndresMolares/Hybrid-ML-management-HAB-impact}. The dataset was obtained following an official request to INTECMAR which does not allow its distribution.

\bibliographystyle{unsrt}  
\bibliography{references}  

\end{document}